\documentclass[conference]{IEEEtran}

	\usepackage{etoolbox}
\makeatletter
\patchcmd{\@makecaption}
  {\scshape}
  {}
  {}
  {}
\makeatother
  	\usepackage[pdftex]{graphicx}
  	\graphicspath{{../pdf/}{../jpeg/}}
	\DeclareGraphicsExtensions{.pdf,.jpeg,.png}

	\usepackage[cmex10]{amsmath}
	\usepackage{mathabx}
	\usepackage{algorithmic}
	\usepackage{array}
	\usepackage{mdwmath}
	\usepackage{mdwtab}
	\usepackage{eqparbox}
	\usepackage{url}
\usepackage{epstopdf}
\AppendGraphicsExtensions{.tif}

\begin{document}
\title{\LARGE Reconfigurable Co-Processor Architecture with Limited Numerical Precision
 to Accelerate Deep Convolutional Neural Networks}

 \author{\authorblockN{Sasindu Wijeratne, Sandaruwan Jayaweera, Mahesh Dananjaya and Ajith Pasqual}
 \authorblockA{Dept. of Electronic and Telecommunication Engineering, University of Moratuwa, Sri Lanka \\ \{130665u,130254j,110089m\}@uom.lk, pasqual@mrt.ac.lk}}

\maketitle

\begin{abstract}
Convolutional Neural Networks (CNNs) are widely used in deep learning applications, e.g. visual systems, robotics etc. However, existing software solutions are not efficient. Therefore, many hardware accelerators have been proposed optimizing performance, power and resource utilization of the implementation. Amongst existing solutions, Field Programmable Gate Array (FPGA) based architecture provides better cost-energy-performance trade-offs as well as scalability and minimizing development time. In this paper, we present a model-independent reconfigurable co-processing architecture to accelerate CNNs. Our architecture consists of parallel Multiply and Accumulate (MAC) units with caching techniques and interconnection networks to exploit maximum data parallelism. In contrast to existing solutions, we introduce limited precision 32 bit Q-format fixed point quantization for arithmetic representations and operations. As a result, our architecture achieved significant reduction in resource utilization with competitive accuracy. Furthermore, we developed an assembly-type microinstructions to access the co-processing fabric to manage layer-wise parallelism, thereby making re-use of limited resources. Finally, we have tested our architecture up to 9x9 kernel size on Xilinx Virtex 7 FPGA, achieving a throughput of up to 226.2 GOp/S for 3x3 kernel size.

\end{abstract}

\IEEEoverridecommandlockouts
\begin{keywords}
CNN, Reconfigurable Co-Processor, High-Throughput Architecture, Hardware Acceleration, Programmable Processing Fabric, Q-Point Fixed Precision
\end{keywords}

\IEEEpeerreviewmaketitle



\section{Introduction}
Convolutional Neural Networks (CNN or ConvNet) are perhaps the most widely used neural network model for Deep Learning (DL) applications, e.g., image classification, speech recognition, language processing etc. As an example, LeCun et el. in \cite{726791}\cite{4270182} and Hinton et el. in \cite{Krizhevsky:2012:ICD:2999134.2999257} provided details of such applications with record accuracy. ConvNets have simple kernel based computational structures which significantly reduce their computational time and resources. As a result, ConvNets are algorithmically simpler and more accurate compared to other neural network models \cite{Krizhevsky:2012:ICD:2999134.2999257}. 

CNNs are very computationally intensive and the convolution layers account for the largest part by far. This is because convolution require a large number of multiply-accumulation operations. Computational complexity of the ConvNet is also increasing with today's complex learning models, representations and dimensionality of data. Therefore, it is increasingly challenging to train DL models and infer based on them. However, existing software solutions are not efficient. Therefore, many accelerators have been proposed over the years to efficiently carry-out CNNs on hardware. Particularly, number of FPGA based hardware accelerators \cite{Zhang:2015:OFA:2684746.2689060}\cite{Suda:2016:TOF:2847263.2847276}\cite{Qiu:2016:GDE:2847263.2847265}\cite{DiCeccoLVCTA16}\cite{zyncnet} have been explored, taking advantage of their reconfigurability, programmability and low power. Among FPGA accelerators, co-processing architectures \cite{DiCeccoLVCTA16}\cite{zyncnet} have significant flexibility and scalability. Also, well designed FPGA architectures can be used to exploit high level of data parallelism.  
Advancements in today's ConvNet accelerators focus on optimizing cost-energy-performance. However, the complexity of deep learning applications have also created a demand for better performance together with high accuracy. In the recent past, single and double precision floating point arithmetic was mostly used. An architecture based on high precision, e.g. floating-point arithmetic, is also relatively resource intensive and power consuming, but provides higher accuracy to the applications. Therefore, number of studies \cite{DBLP:journals/corr/GuptaAGN15}\cite{DBLP:journals/corr/GyselMG16}\cite{10.1007/978-3-319-11179-7_36} have highlighted the importance of precision in ConvNet implementations. Sakr et al. in \cite{pmlr-v70-sakr17a}  and Gupta et al. in \cite{DBLP:journals/corr/GuptaAGN15} have demonstrated competitive results of deep neural networks with limited numerical precision. Also, the precision or representation of numerical values is directly associated with the resource utilization, thus cost. More importantly, with diverse range of deep learning applications, the cost-performance-energy-accuracy trade-off has leveraged its importance. Therefore, it is important to achieve high performance and accuracy with limited and reduced precision. However, most of the previous implementations were heavily depending on the floating point, either single precision or double precision. This approach might not be best suited for limited resource environments, e.g. embedded applications.

In this paper, we present a novel reconfigurable co-processing architecture to accelerate ConvNets. We have also introduced the Q-format fixed point quantization arithmetic to reduce the resource utilization of the FPGA hardware while maintaining the accuracy at a competitive level. This approach significantly reduces the resource utilization and processing time, enabling the use of this architecture for range of embedded applications. Also, proposed architecture is highly model independent and efficient, but reconfigurable and programmable. In addition, we introduce CISC like microinstructions to control the hardware operations at run time which is used to gain layer-wise parallelism reusing limited resources. 

Section II briefly describes the background of our research including Q-point arithmetics. In section III, we explain the design space exploration for our architecture including data parallelism, caching and pipelined operations. Finally, in Section IV, details of the implementation along with results obtained for this new architecture are provided with a comparison. In our implementation, we used ImageNet \cite{imagenet_cvpr09} dataset of 2012 ILSVRC competition. Also, for the comparison purpose we used AlexNet \cite{Krizhevsky:2012:ICD:2999134.2999257} and ZyncNet \cite{zyncnet}. Section V provides the conclusion with possible future developments.

 

\section{Background}
A ConvNet is a multi-layer feed forward neural network with convolution filters and nonlinearity \cite{726791}\cite{4270182}. Over the past few years, many different CNN architectures have been proposed to address the efficiency and accuracy of various learning tasks. As a subsequent outcome, different approaches like LeNet \cite{726791}, AlexNet \cite{Krizhevsky:2012:ICD:2999134.2999257}, VGG \cite{DBLP:journals/corr/SimonyanZ14a}, GoogLeNet \cite{DBLP:journals/corr/SzegedyLJSRAEVR14}, ResNet \cite{DBLP:journals/corr/HeZRS15}, SqueezeNet \cite{DBLP:journals/corr/IandolaMAHDK16}, have been proposed to accelerate the learning and inferencing with better performance. In general, ConvNet architectures consist of several layers, i.e. convolution layers, activation and pooling layers \cite{Krizhevsky:2012:ICD:2999134.2999257} organized in different configurations.

In CNNs, the output feature map is obtained by following steps. For each ConvNet filter, input feature map of length $l_{in}$, width $w_{in}$ and depth $D_{in}$ is convoluted with a shifting $k \times k $ large kernel with same depth of $D_{in}$. Then convoluted data will be passed through activation function. The sigmoid, tanh and ReLu 
are commonly used as activation functions for ConvNet. In order to reduce the spatial size, computational complexity and number of parameters pooling layers are used in between successive Convolutional layers. 
 
The mathematical representation of convolution layer and activation function is shown in equation \ref{eq:1}.
\begin{equation} \label{eq:1}
\vartheta_{t} = \textbf{F}(\sum_{r=0}^{D_{in}-1}\sum_{i=0}^{k-1}\sum_{j=0}^{k-1} (weight_{(i,j,r,t)} \times local\_input_{(i,j,r,t)}) + b_t)
\end{equation}

Here, $\vartheta_{t}$ ,  $D_{in}$, \textit{k}, \textbf{F} and  \textit{b} represent the corresponding output feature, depth of input feature, kernel length, activation function and bias respectively.

DiCecco et al. in \cite{DiCeccoLVCTA16} proposed an end-to-end FPGA accelerated co-processing framework for Caffe CNN in which an FPGA layer can be used as a co-processor alongside other layers running on a host processor. But, in such an approach, memory becomes a bottleneck when the size of the layers and parameters are increasing. Therefore, users have the flexibility to decide which part of the program should be running on FPGA, either a specific portion or the complete program. Therefore, flexibility and programmability is one of the key areas of focus in FPGA accelerated architectures. The problems was addressed by some techniques such as an instruction set to program CNN hardware \cite{Krizhevsky:2012:ICD:2999134.2999257}.  

Several other research explored the implementation of CNNs on FPGAs \cite{Zhang:2015:OFA:2684746.2689060} \cite{Suda:2016:TOF:2847263.2847276}\cite{Qiu:2016:GDE:2847263.2847265} to make use of the low power, reconfigurability and programmability. 
A more comprehensive and state-of-the-art FPGA based accelerator was proposed by David Gschwend in his thesis \cite{zyncnet}, ZyncNet-An FPGA-Accelerated Embedded Convolutional Neural Network. This architecture demonstrated competitive results in performance, accuracy and memory management compared to existing architectures \cite{Krizhevsky:2012:ICD:2999134.2999257}\cite{DBLP:journals/corr/SimonyanZ14a}\cite{DBLP:journals/corr/SzegedyLJSRAEVR14}\cite{DBLP:journals/corr/HeZRS15}\cite{DBLP:journals/corr/IandolaMAHDK16}. Even though their approach produced highly competitive results, the design used model specific approach and High Level Synthesis (HLS). The authors also used single precision floating point for representations and arithmetic operations, causing high resource utilization.

The main reason behind application specific designs is to support parallelism \cite{DBLP:journals/corr/Krizhevsky14}\cite{DiCeccoLVCTA16}. According to existing techniques, data parallelism is highly exploited on custom hardware designs alongside model parallelism, layer parallelism and pipeline parallelism. One or more of these techniques can be found in almost all existing accelerators, e.g. ZyncNet \cite{zyncnet}. In particular, inherently parallel pixel operations can be carried out concurrently when CNNs are used in image processing applications. Pipeline parallelism is applied when operating different dependent steps of operations concurrently on parallel threads which is well suited for feed forward computations of CNNs. Most accelerators have used these techniques in the past.

The Q-Point representation is a fixed point format where the number of fractional bits and integer bits are specified prior to the usage. Depending on the number of bits in its representation, Q Format limits the range of numbers it can represent with an acceptable degree of accuracy \cite{DBLP:journals/corr/CourbariauxBD14}. The Q-Format number is represented as $Q_{(n-m-1,m)}$ where n is the total number of bits, m is fractional bits and single sign bit. Overflow is avoided by using the proper number of fractional and integer bits depending on the weights and the input feature data. This simple representation of numbers makes the arithmetic operators for Q-Point representation hardware efficient. 
Furthermore, fixed point Q-Format numerical arithmetic significantly reduces resource utilization and power consumption. But, it has a trade off with numerical precision and accuracy. However, some recent research suggests that numerical precision might not be the case for CNNs \cite{pmlr-v70-sakr17a}\cite{DBLP:journals/corr/GuptaAGN15}\cite{DBLP:journals/corr/CourbariauxBD14}.

\section{Architecture}

The proposed architecture is directly focusing on Field Programmable Gate Arrays (FPGAs). This accelerator architecture includes reconfigurable-parameters which provides the ability of re-reconfigurability, depends on the CNN architecture which run on top of the accelerator. The reconfigurable-parameters which used in this system are as follows:

\begin{itemize}
  \item The depth of input features ($D_{in}$)
  \item Filter Size ($S$)
  \item Number of filters ($N$)
  \item Pooling Filter Size ($P$)
  \item Activation Function Selection ($Sel_{AF}$)
  \item Pooling Layer Configuration ($Conf_{p}$)
\end{itemize}

The Processing Fabric (PF) is the core logic processing area of the design. Moreover, the it is highly parallel and can be reused for multiple layers by accessing the parameters in each layer from the internal instruction memory.

\begin{figure}[t] 
\centering
\includegraphics[width=3.5in]{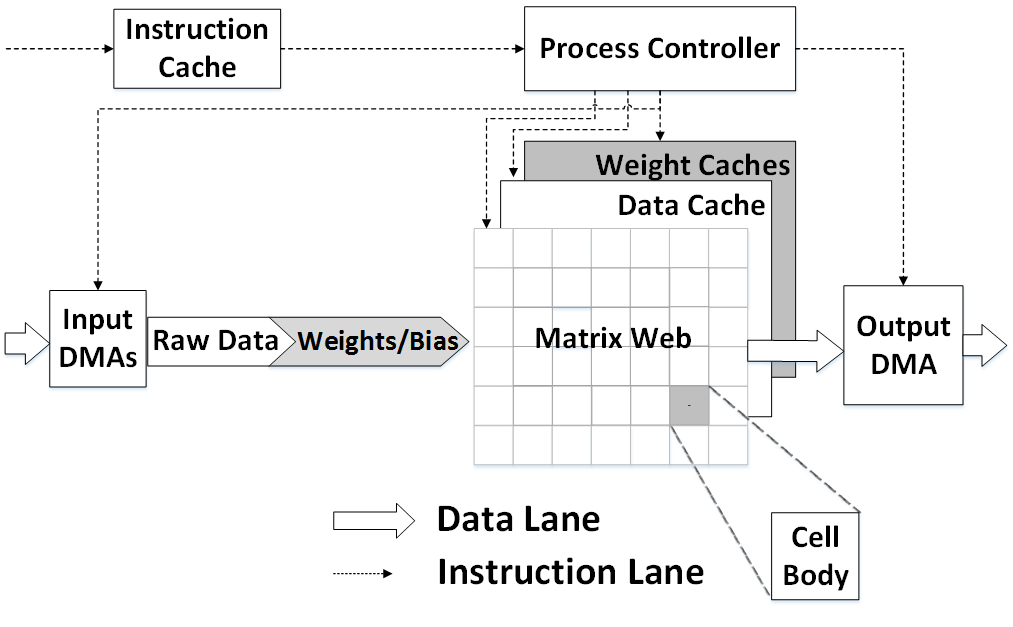}
\caption{The Overall Conceptual Architecture}
\label{overall_1}
\end{figure}
The parallelism of independent operations is the state-of-the-art for accelerating neural networks. In this architecture, we have identified such independent processes and implemented them in parallel in order to exploit a high level of data parallelism. However, such a design is hardware intensive and consumes a large amount of power. This problem is reduced by reusing limited resources with layer wise parallelism. Also, this architecture use parallel multi-channels to accelerate processing feature data in depth ($D_{in}$). Before running a CNN on top of this processing fabric, $D_{in}$ should be set to input feature depth of the processing CNN architecture.

The Q-Format fixed point arithmetic is used in the architecture which consists of less hardware intensive and less time consuming arithmetic operations. Therefore, it has advanced the overall architecture as well.



Instruction set architecture gives the ability to process different input features sizes and enable zero padding. This gives the architecture more flexibility to run different CNN architectures on top of this hardware acceleration platform.

\begin{figure}[t]
\centering
\includegraphics[width=3.5in]{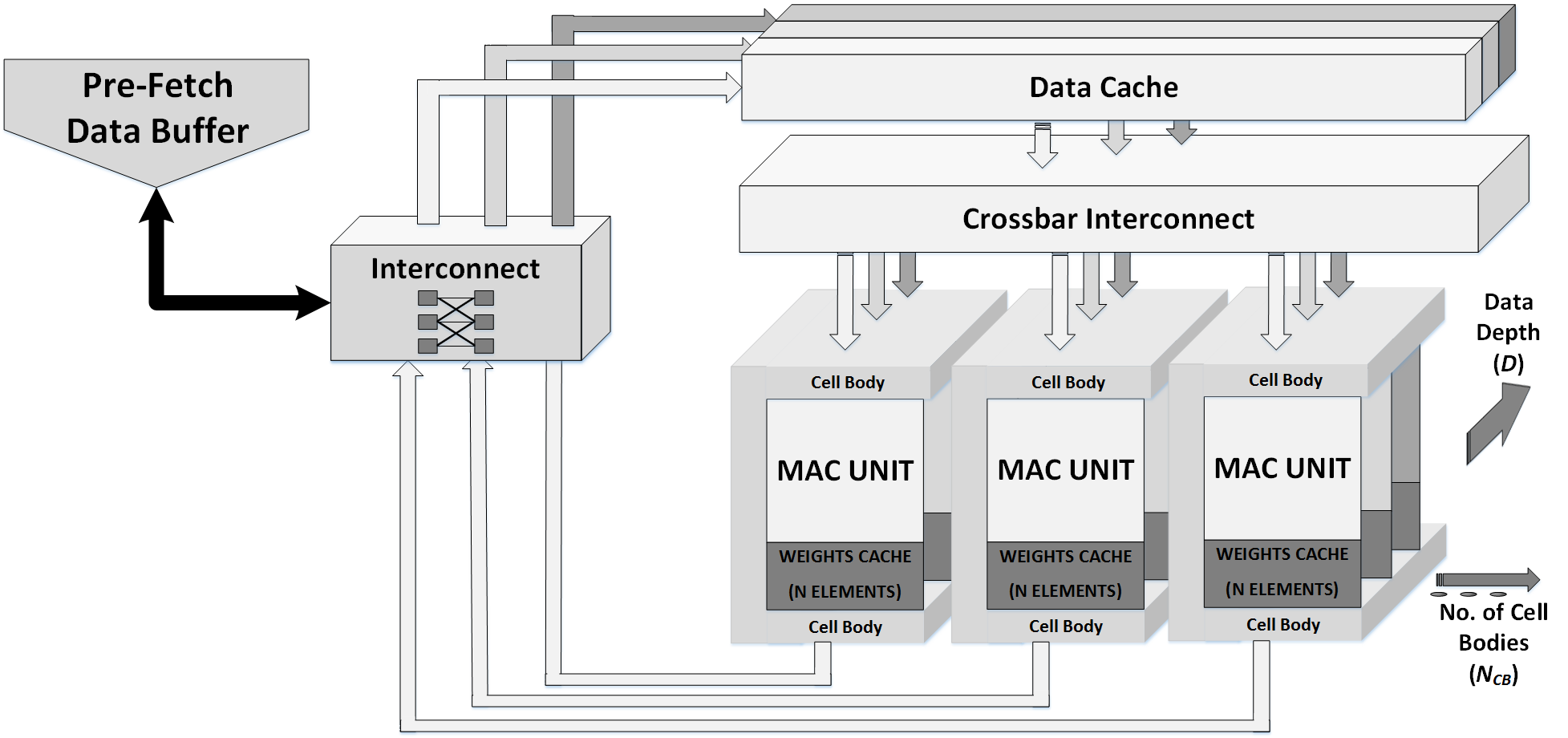}
\caption{The internal structure of Matrix Web}
\label{synapse_1}
\end{figure}

The conceptual design of the architecture is shown in figure \ref{overall_1}. It contains separate data flow path and instruction flow path as in Harvard Architecture. As shown in the figure \ref{overall_1}, there are 2 major units in this system. They are Process Controller (PC) and the Matrix Web (MW). The main responsibility of the Process Controller is to fetch instruction and execute instructions. The instruction execution is mainly focused on memory addressing. The MW consists of arithmetic and logic units which are the main functional elements of the processing architecture. The input feature data, weights and biases are cached within the Matrix Web. This caching system and interconnection is extensively shown in figure \ref{synapse_1}. Moreover, using Direct Memory Access controllers (DMA), data is transfered as bulk between main memory and processing fabric in order to reduce the total number of execution Instructions and execution time per layer. After instructions are fetched into the system, first, the weights and bias for each kernel is loaded into dedicated caches through data lines. Input data is then fetched and processed in the MW and processed. Then processed data is passed on to data buffer and finally, output data is moved to main memory through DMA transactions.

The Matrix Web (MW) is the major arithmetic and logic unit in the system. Figure \ref{synapse_1} shows a detailed structure of MW. The size of the MW depends on the number of filters ($N$), filter size ($S$) and the depth of the input features($D_{in}$) which are reconfigurable parameters to the processing fabric. 

\begin{figure}[b]
\raggedleft
\includegraphics[width=3in]{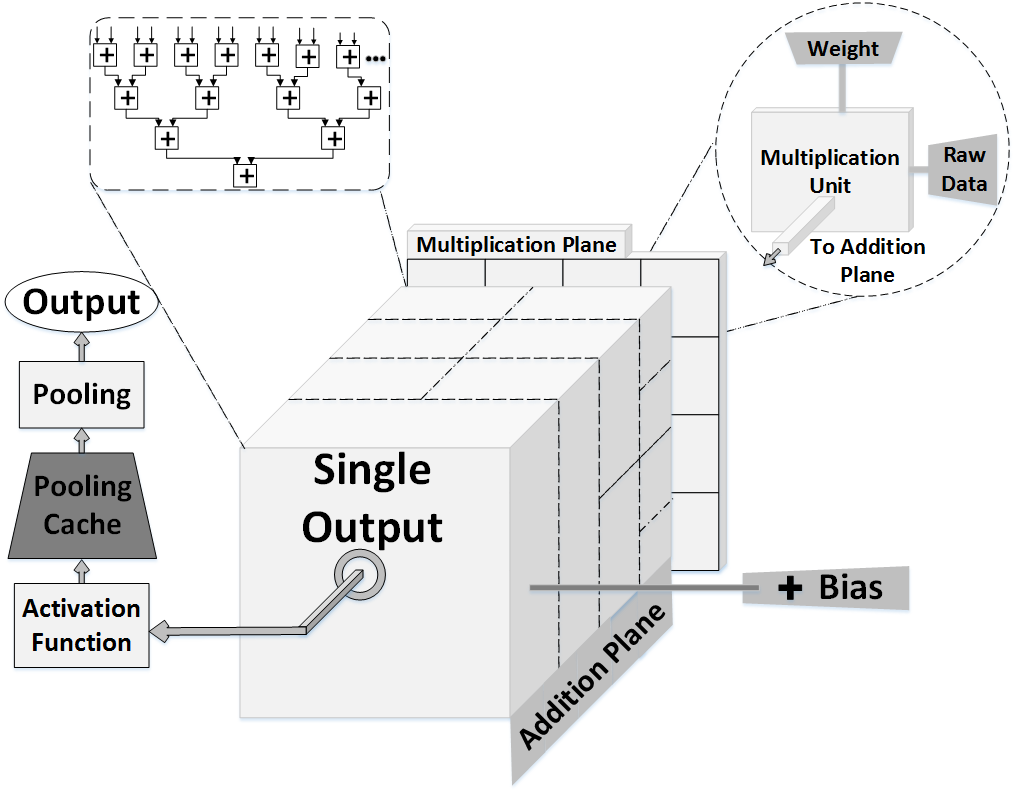}
\caption{The Cell Body Unit}
\label{Adder_2}
\end{figure}

As shown in the figure \ref{overall_1}, Cell Body Units (CBU) are the basic processing elements to the processing fabric. It simulates the process of convolution filter. The CBU calculates the output feature data according to the equation \ref{eq:1}. Each CBU has internal weight caches, internal MAC units, an activation layer and a pooling layer as shown in figure \ref{Adder_2}. pooling layer can be enabled while configuring processing fabric by setting the $Conf_p$ reconfigurable parameter. $Conf_p$ parameter expects width of pooling kernel. Activation function to the CBU unit is also a reconfigurable parameter ($Sel_{AF}$). For each MAC unit there is a dedicated Weight cache. The weight caches are filled before feeding the input data. The input data is first buffered into a pre-fetch data buffer and pass into the caches through a simple interconnect. Each CBU is synchronized to operate in the same input data in same clock cycle. Therefore, data cache is shared between each CBU, using a Crossbar interconnection. The output data of different kernels can be processed massively parallel by using more CBUs. The final calculated data is buffered and forwarded in to the main memory as shown in the figure \ref{Adder_2}. This massively parallel architecture reduces the input feature data re-usage. The number of MAC Units per Cell Body depend on the depth of input features and the number of CBUs depend on number of filters that can be parallel processed depend on the hardware availability of FPGA as shown in figure \ref{synapse_1}. The instruction set gives the flexibility to use number of parallel CBUs in this manner. Using high number of parallel CBUs we can minimize the input feature data re-usage. 

In CBU after Weight cache, bias cache and input feature cache is filled with data, they forwarded MAC units according to the instructions from PC. In MAC units, input feature is multiplied and added together. In the first layer of Addition Plane, $2^{(k-1)} \times D_{in}$ addition units are implemented. If the kernel size is not a power of 2, the unoccupied Addition Units (AU) are padded zero. This process is pipelined in order to increase the throughput.  Such a implementation makes the system work at a high frequency. The number of Multipliers and Adders depends on $S$ and $D_{in}$, the reconfigurable parameters to the Processing Fabric.



After processing through MACs, results of the MAC units are forwarded to BA. The number of MAC units connected to the BA depends on ($D_{in}$) as shown in \ref{Adder_2}. They are connected using series of adder layers as in MACs. After the Bias adder, the result is forwarded to AF. The activation function is enabled and selected in the configuration level, therefore it is a reconfigurable parameter. Activation functions include Sigmoid, tanh, ReLU and Max. After AF data is stored in a temporary cache call Pooling cache. Thereafter the data is proceed to the pooling layer. The size of Pooling cache depends on the pooling kernel size. In this implementation MAX pooling is used. The scalability of the Pooling layer is maintained by using the width of pooling kernel ($Conf_{p}$). Making pooling kernel size to 1 is similar to deactivating the pooling layer. Finally, processed data is passed to the data buffer.

\begin{multline}
\text{No. of mul. units per MAC Unit = } k^2
\end{multline}
\begin{multline}
 \text{No. of mul. units per Cell Body = } D_{in} \times k^2
\end{multline}
\begin{multline}
\text{No. of adders per Cell Body = } D_{in}\times\sum\limits_{r = 0}^{\lceil (2 \times log_2(k)) \rceil} 2^{r} \\ + \sum\limits_{p = 0}^{\lceil (log_2(D_{in})) \rceil} 2^{p} 
\end{multline}

Here, $k$ and $D_{in}$ stand for filter width and input feature data depth.

The Process Controller (PC) is responsible for instruction execution. The scalability in the configuration level is achieved by using the scalable instruction set which scales with the number of kernels in the system, number of weights per kernel and depth of the input. The Instructions that fetch into the PC follows the CISC (complex instruction set computer) instruction type. In the Process Controller unit, there are 2 basic types if instructions as shown below.
\\\\\textbf{MatrixWeb Control Instructions:}  feature width, feature depth, Stride Length, Zero Padding enable, Convolve. \\\\\textbf{Memory Control Instructions:}  Address space of Input features, Address space of Weights, Address space of Biases, Address space of Outputs.\\

 The memory locations of weights and biases related to each layer are injected into the processing fabric with Memory Control Instructions. MatrixWeb Control Instructions Configure the Matrix Web according the CNN network that is to be processed in the accelerator. The instruction set gives the flexibility to use dynamic number of parallel Cell Bodies which depend on the hardware limitation of FPGA. 
 
\begin{figure}[ht!] 
\centering
\includegraphics[width=3.5in]{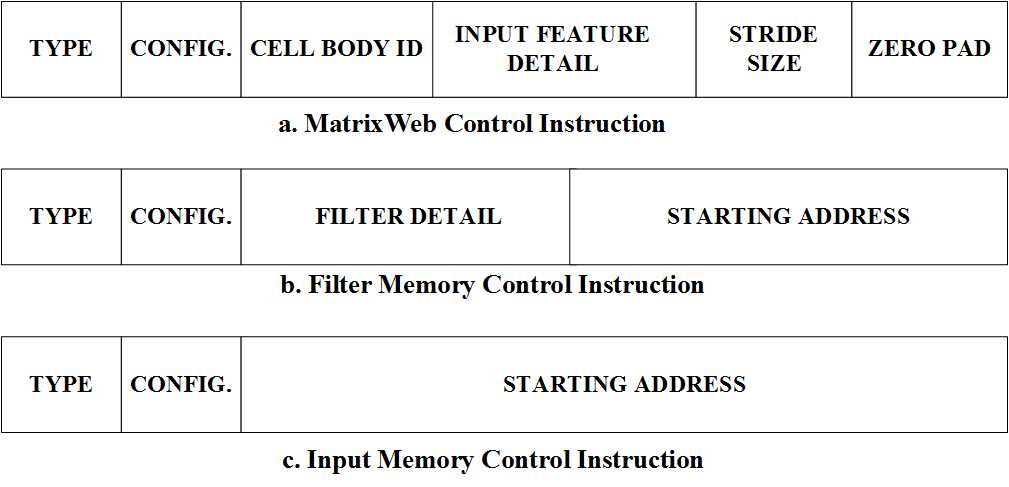}
\caption{Basic Instructions of the Architecture}
\label{ins}
\end{figure}

As shown in figure \ref{ins}, MatrixWeb Control Instructions and Memory Control Instructions are the main type of instructions for our system. It contains several fields for each type. In both instruction types, the TYPE field is in common. As the name suggest, it is used to identify the type of the instruction, MatrixWeb Control Instruction or Memory Control Instruction.

In MatrixWeb, Control Instruction CONFIG field is responsible for giving the system the instruction of starting convolution, clearing (or flushing) the weights and the biases that are cached within MAC units and stop convolution. The Input Feature Detail (IFD) field is required to identify the depth and width of the input feature data. It gives the system the flexibility to process different sizes of input data. The Stride Len. (SL) specifies the stride with which the filter is slided. The Zero-Pad (ZP) field is used to zero pad the input features using internal logic which reduces the input feature size. In addition, enabling zero padding through instruction will decrease the requirement of input feature space. In order to configure them, MatrixWeb Control Instructions are released to each Cell Body Unit.

Filter Memory Control Instructions are used to point out the memory space of Weights, the memory space biases and the Memory space of Output feature for each corresponding Cell Body. There are 3 Filter Memory Control Instructions for each cell body. There is one Input Memory Control instruction for the system for single input feature data set.

 The total number of cycles that is needed to load Weights and biases depend on the number of kernels, the depth of the input data set, number of weights and bias per kernel. The number of  cycles that is needed to process input feature data is directly correspond to the width and depth of the data set.

\begin{equation}
C = \gamma + (D_{in} + 2) \times \gamma + 1
\end{equation}


C is the number of cycles to fetch all (total) instructions. Here $\gamma$ and $D_{in}$ represents the number of cell bodies and input feature data depth.


Data is transferred between the main memory of the processing system and hardware accelerator unit using DMAs over PCIe. Multiple channels are used in order to minimize the data traffic.


\section{Performance Evaluation}

We developed a software model identical to our hardware architecture using C/C++. It is capable of handling fixed point as well as single and double precision floating points. Furthermore, the architecture was implemented using Verilog which was later used with software simulation to verify results. Both software and hardware implementations are combined together with System Verilog and Direct Programming Interface (DPI) to create hardware simulation. The final design was synthesized and implemented on Xilinx Virtex-7 FPGA XC7VX485T using Vivado 2015.4 software. Also, our co-processor was designed to be connected with host machine via PCIe interface. 

With the provided reconfigurability and programmability, our framework capable of handling different CNN architectures, e.g. AlexNet, SqueezeNet, ZyncNet etc. But, for the accuracy comparison, in this section, we present results based on accelerating two different CNN architectures on two base designs.

\begin{table}[ht!] 
\centering
\label{accuracy-compare}
\caption{Comparison of proposed architecture to existing CNN architectures. 'Layers' is the number of convolution layers. Also, MACCs, Parameters and Activations are in millions}
\begin{tabular}{c c c c c c}\hline
\textbf{} & \textbf{Layers} & \textbf{MACCs} & \textbf{Params} & \textbf{Activs.} & \textbf{Top-5 Error} \\\hline\hline
Case B* & 18 & 530 & 2.5 & 8.8 & 15.7\% \\\hline
ZyncNet & 18 & 530 & 2.5 & 8.8 & 15.4\% \\\hline
Case A* & 5 & 1140 & 62.4 & 2.4 & 20.3\% \\\hline
AlexNet & 5 & 1140 & 62.4 & 2.4 & 19.7\% \\\hline
VGG-16 & 16 & 15470 & 138.3 & 29.0 & 8.1\% \\\hline
GoogLeNet & 22 & 1600 & 7.0 & 10.4 & 9.2\% \\\hline
ResNet-50 & 50 & 3870 & 25.6 & 46.9 & 7.0\% \\\hline
SqueezeNet & 18 & 860 & 1.2 & 12.7 & 19.7\% \\\hline
\end{tabular}
\end{table}

 In case A,  AlexNet CNN architecture and in case B, ZyncNet CNN architecture are processed on top of our processing fabric
by setting up reconfigurable parameters accordingly.



For each CNN architecture, we use both software and hardware simulation to train the model based on ImageNet dataset. Thereafter, we use the same setup with trained weights to predict on ImageNet validation data set, particularly with $Q_{(16,15)}$, Q-point representation in hardware. Since, our software simulation is capable of handling both fixed and floating precision, we calculated the absolute difference (error) for each data point and operation. Table I provides a details of final top-5 accuracy obtained from the results. As shown in the table, final accuracy has dropped, particularly with 0.6\% in the Case A and 0.3\% in the Case B when using 32 bit fixed precision which is a tolerable amount. 

\begin{table*}[ht!] 
\centering
\label{table-compare-fpga}
\caption{Existing FPGA based CNN compared to proposed architecture}
\begin{tabular}{|c|c|c|c|c|c|c|}\hline
\textbf{Metric} & \textbf{Zhang \cite{Zhang:2015:OFA:2684746.2689060}} & \textbf{Suda \cite{Suda:2016:TOF:2847263.2847276}} & \textbf{Qui \cite{Qiu:2016:GDE:2847263.2847265}} & \textbf{DiCecco \cite{DiCeccoLVCTA16}} & \textbf{ZynqNet \cite{zyncnet}} & \textbf{Proposed} \\\hline\hline
Frequency(MHz) & 100 & 120 & 150 & 200 & 200 & 200 \\\hline
Precision & 32 bit float & 8/16 bit fixed & 16 bit fixed & 32 bit float & 32 bit float & 32 bit fixed \\\hline
FPGA Version & Virtex 7 VX485T & Stratix-V GSD8 & Zynq XC7Z045 & Virtex 7 XC7VX690T-2 & Zynq XC7Z045 & Virtex 7 XC7VX485T-2 \\\hline
DSP Utilization & 2,240 & (Not specified) & 780 & 1,307 & 739 & 576 \\\hline
Host Connection & on-chip & PCIe & on-chip & PCIe & on-chip & PCIe \\\hline
GFLOPS/GOPS & 61.62 & 136.5 & 187.8 & 50 & (Not Specified) & 226.2 \\\hline
\end{tabular}
\end{table*}


\begin{figure}[ht!]
\centering
\includegraphics[width=3.5in]{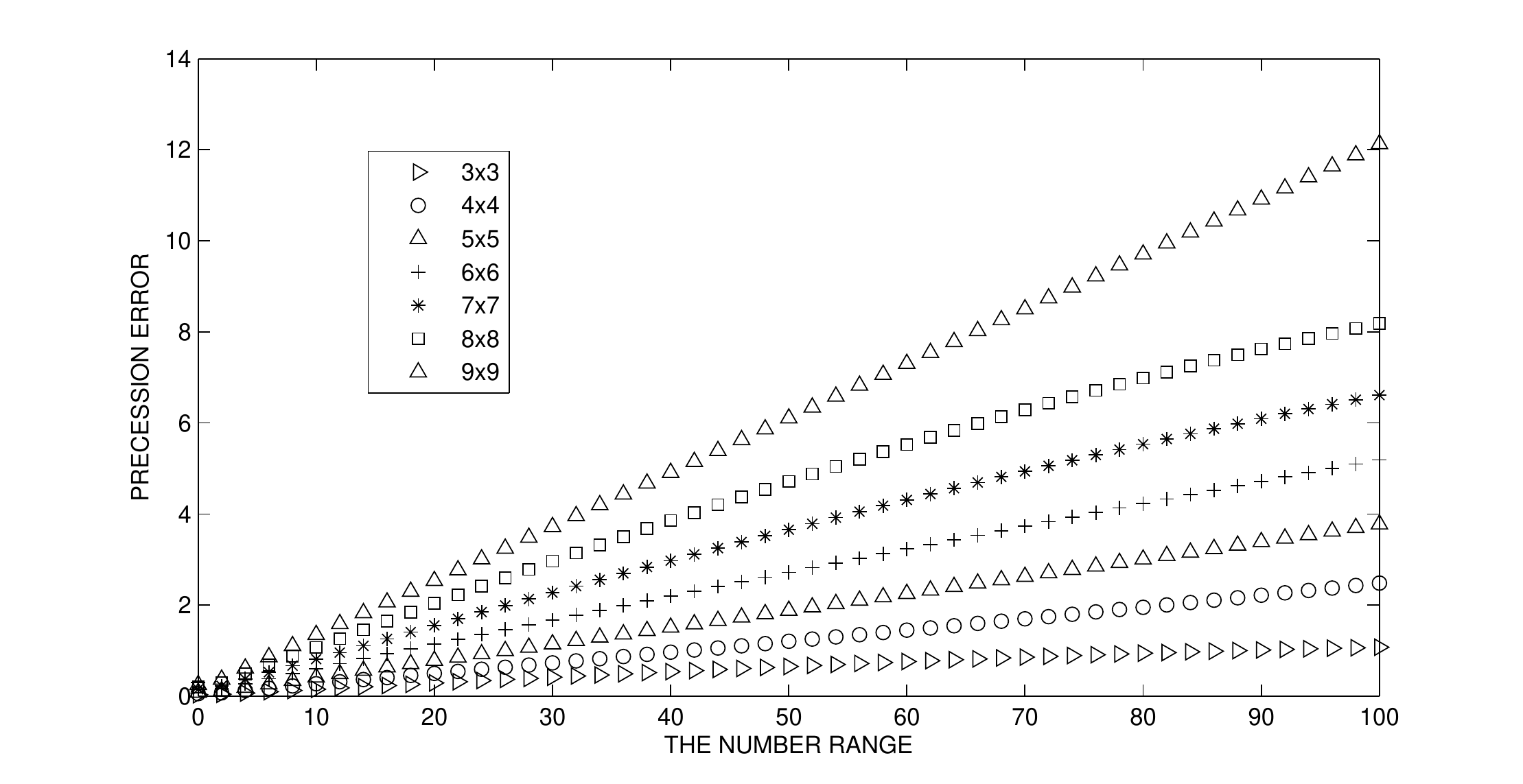}
\caption{Average error as a function of input values for different kernel width}
\label{error_1}
\end{figure}

\begin{figure}[ht!] 
\centering
\includegraphics[width=3.5in]{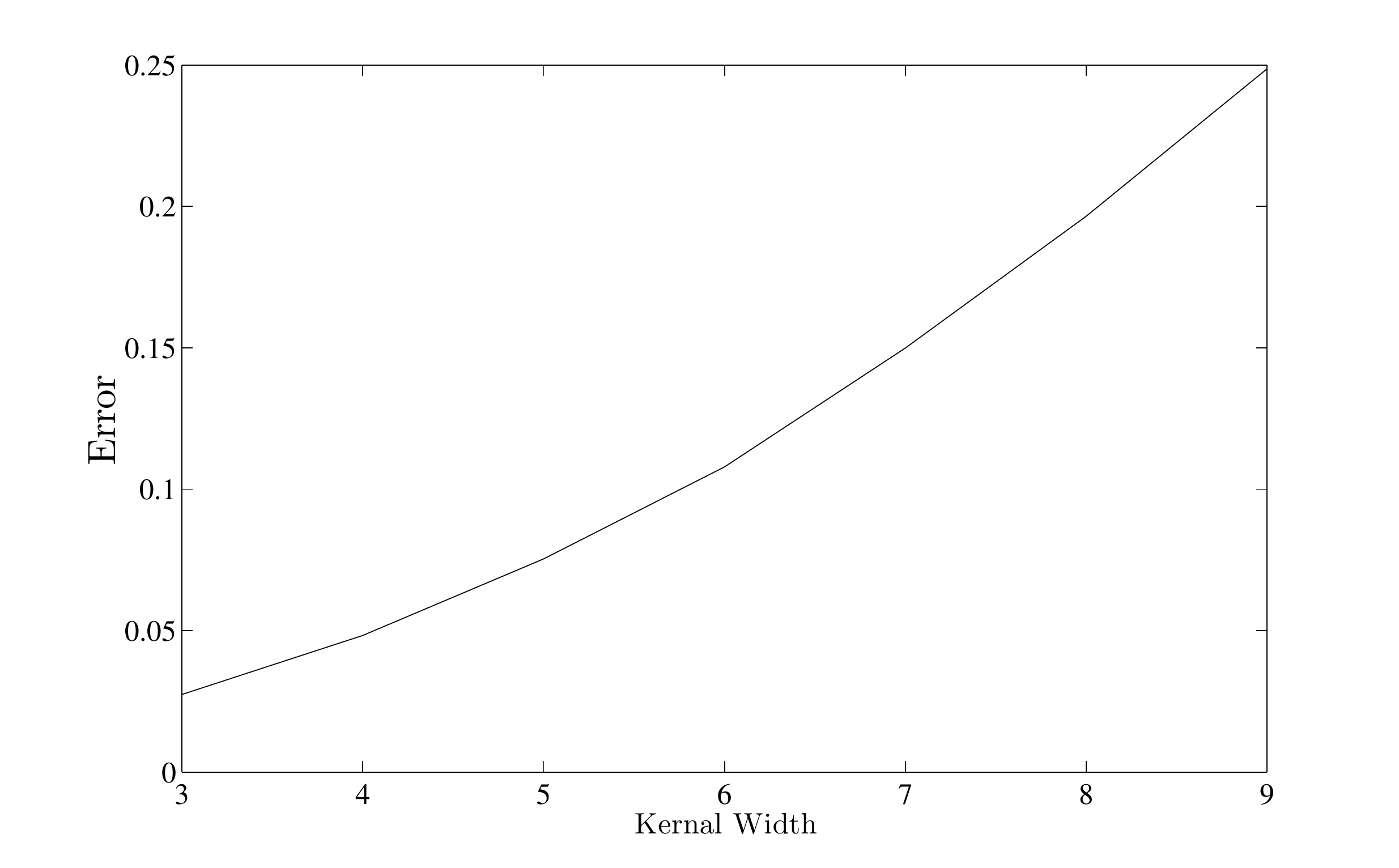}
\caption{Average error as a function of kernel width for input values between 0 and 1}
\label{error_single}
\end{figure}

A similar method is carried-out for finding the accuracy of MAC units. As shown in the figure \ref{error_single}, the accuracy and its variation with respect to different input value ranges and different kernel sizes is  presented and evaluated within this comparison. Since the CNNs often provide only a certain range of input values, it is easy to show that the proposed neural network is sufficiently capable of providing an accurate output for the requirement.

In this performance evaluation $Q_{(16,15)}$, Q-point representation is used. Changing the kernel size and input data values, accumulated error readings have been obtained. These accuracy readings, the error values have been plotted as functions of kernel size and input value.

The test is carried out for different value ranges and different kernel sizes as shown in figure \ref{error_1}. The error for each constant kernel size increases exponentially with the input data values. As in figure \ref{error_1}, the average error for input values 0 to 50 stay well below 0.1 for any kernel width from 3 to 9.

A kernel width of 3 is used as the initial parameter value and is increased up to 9 while maintaining the input values in the range 0 to 1. The error shows an exponential increase with the increase of kernel size as shown in figure \ref{error_single}. Still, the error values are extremely small compared to the input values even for a kernel width of 9. 
Results from Figure \ref{error_single} and Figure \ref{error_1} explains how accuracy drops when the model is getting bigger. Increasing kernel sizes as well as parameters can significantly change accuracy value. It causes much larger model in case A to drop its accuracy more than the much smaller case B model. However, obtained accuracy drops, i.e. 0.6\% and 0.3\% for A and B respectively, lie within acceptable level.

CNN architectures have different requirements. As an example, AlexNet needs up to 11x11 convolutions whereas ZyncNet CNN [16] uses only up to 3x3 convolutions. But for the performance and resource comparison, we used our Case B (with 3x3 kernel size) hardware design and ZyncNet hardware. In this comparison, we use 3x3 kernel size with 16 Cell Body Units. 


\begin{table}[ht!] 
\centering
\caption{Resource Utilization of a single Cell Body Unit (CBU) with DSP for input data depth ($D_{in}$) = 1}
\label{table-dsp-single}
\begin{tabular}{|c|c|c|c|}\hline
\textbf{Kernel Size} & \textbf{LUT} & \textbf{FF} & \textbf{DSP}\\\hline\hline
3X3 & 2416 & 1299 & 36\\\hline
4X4 & 3374 & 2013 & 64\\\hline
5X5 & 5683 & 3157 & 100\\\hline
6X6 & 7913 & 4433 & 144 \\\hline
7X7 & 10536 & 5978 & 196 \\\hline
8X8 & 13143 & 7587 & 256  \\\hline
9X9 & 17338 & 9784 & 324 \\\hline
\end{tabular}
\end{table}
Table \ref{table-dsp-single} illustrates the resource utilization of a single Cell Body Unit with input data depth ($D_{in}$) = 1 whereas  Table \ref{table-dsp-multi} provides the resource utilization of a single Cell Body Unit with input data depth ($D_{in}$) = 3. 

\begin{table}[ht!] 
\centering
\caption{Resource Utilization of a single Cell Body Unit (CBU) with DSP for input data depth ($D_{in}$) = 3}
\label{table-dsp-multi}
\begin{tabular}{|c|c|c|c|}\hline
\textbf{Kernel Size} & \textbf{LUT} & \textbf{FF} & \textbf{DSP} \\\hline\hline
3X3 & 6721 & 3159 & 108 \\\hline
4X4 & 9360 & 4857 & 192  \\\hline
5X5 & 15039 & 7713 & 300 \\\hline
6X6 & 22169 & 10837 & 432 \\\hline
7X7 & 29487 & 14626 & 588 \\\hline
8X8 & 36812 & 18503 & 768 \\\hline
9X9 & 48684  & 23992 & 972 \\\hline
\end{tabular}
\end{table}


To compare with ZyncNet, we configured our co-processor with 16 Cell Body Units, 3x3 kernel size and $D_{in}$=1. We obtained results as shown in Table V\label{comparison-zynq}. According to the results, our approach shows significant reduction (about 22\%) in the DSP utilization. Moreover, our proposed architecture provides 226.2 GOPS at frequency of 200MHz. Compared to 16 bit fixed precision implementations like \cite{Qiu:2016:GDE:2847263.2847265}, our approach with 32 bit fixed precision has produce 22.5\% more GOPS. More details and a comparison can be found in the Table II.

\begin{table}[ht!] 
\centering
\label{comparison-zynq}
\caption{Resource Comparison between ZynqNet and Proposed}
\begin{tabular}{|c|c|c|c|}\hline
\textbf{Method} & \textbf{LUT} & \textbf{FF} & \textbf{DSP} \\\hline\hline
ZynqNet \cite{zyncnet} & 154K & 137K & 739 \\\hline
Proposed & 117K & 21K & 576 \\\hline
\end{tabular}
\end{table}
According to the Table V\label{comparison-zynq}, our approach shows less resources utilization than previous approach. This is a direct result of using fixed precision and developing a ground-up architecture avoiding resource intensive High Level Synthesis (HLS).
\section{Conclusion}
In this paper, we presented a novel co-processing architecture for Convolutional Neural Networks (CNNs), suitable for reconfigurable devices such as FPGAs. It is developed as a co-processor to accelerate existing software frameworks and CNNs. As results showed, our approach is more scalable and high throughput. Targeted for embedded applications, the architecture demonstrates significant design flexibility. The programmable processing fabric can be reused for multiple layers by accessing and storing the hyper-parameters in each layer. The Instruction Set Architecture (ISA) is capable of handling convolutional layer operations. Moreover, our common instruction set can be used to optimize different high level programs into FPGA. Also, fixed point Q-Format precision provided results that are sufficient for CNN computation with reduced time and resource consumption providing less complexity to the hardware. However, our result shows that large models with high number of parameters have significant accuracy deviation compared to small models. Therefore, fixed precision can be efficiently used to accelerate more compressed modes in hardware. It is also possible to further accelerate the overall system by taking other layers including the Pooling layer the hardware layer. We also showed that 32 bit fixed precision has significantly reduced the operation time and resource utilization while maintaining both accuracy and throughput at higher level. These results encourage future advancements with further reduction in precision, e.g. 16 and 8 bit in fixed point, in hardware accelerators.




\bibliographystyle{IEEEtran}
\bibliography{IEEEabrv,biblio_traps_dynamics}

\end{document}